\documentclass[sigplan,  10pt]{acmart}
\usepackage{enumitem}
\usepackage{subcaption}
\usepackage[algo2e,ruled,linesnumbered]{algorithm2e}
\usepackage{caption}
\usepackage{subcaption}

\author{Vani Nagarajan}
\affiliation{
    \department{School of Electrical and Computer Engineering}
    \institution{Purdue University}
    \city{West Lafayette}
    \state{IN}
    \country{USA}}
\email{nagara16@purdue.edu}

\author{Durga Mandarapu}
\affiliation{
    \department{Department of Computer Science}
    \institution{Purdue University}
    \city{West Lafayette}
    \state{IN}
    \country{USA}}
\email{dmandara@purdue.edu}

\author{Milind Kulkarni}
\affiliation{
    \department{School of Electrical and Computer Engineering}
    \institution{Purdue University}
    \city{West Lafayette}
    \state{IN}
    \country{USA}}
\email{milind@purdue.edu}

\usepackage{booktabs}   
\usepackage{subcaption} 

\usepackage{amsmath}    
\usepackage{caption}
\usepackage{listings}
\usepackage{xcolor}
\usepackage{graphicx}
\usepackage{xspace}
\usepackage{algorithm}
\usepackage{algpseudocode}
\usepackage{multirow}
\acmYear{2023}\copyrightyear{2023}
\setcopyright{rightsretained}
\acmConference[ICS '23]{International Conference on Supercomputing}{June 21--23, 2023}{Orlando, FL, USA}
\acmBooktitle{International Conference on Supercomputing (ICS '23), June 21--23, 2023, Orlando, FL, USA}
\acmPrice{}
\acmDOI{10.1145/3577193.3593738}
\acmISBN{979-8-4007-0056-9/23/06}

\newcommand{\etal}{\textit{ et. al. }}

\title{RT-kNNS Unbound: Using RT Cores to Accelerate Unrestricted Neighbor Search}

\begin{document}

\begin{abstract}
The problem of identifying the k-Nearest Neighbors (kNNS) of a point has proven to be very useful both as a standalone application and as a subroutine in larger applications. Given its far-reaching applicability in areas such as machine learning and point clouds, extensive research has gone into leveraging GPU acceleration to solve this problem. Recent work has shown that using Ray Tracing cores in recent GPUs to accelerate kNNS is much more efficient compared to traditional acceleration using shader cores. However, the existing translation of kNNS to a ray tracing problem imposes a constraint on the search space for neighbors. Due to this, we can only use RT cores to accelerate {\em fixed-radius} kNNS, which requires the user to set a search radius {\em a priori} and hence can miss neighbors. In this work, we propose TrueKNN, the first {\em unbounded} RT-accelerated neighbor search. TrueKNN adopts an iterative approach where we incrementally grow the search space until all points have found their {\em k} neighbors. We show that our approach is orders of magnitude faster than existing approaches and can even be used to accelerate fixed-radius neighbor searches.
\end{abstract}



\maketitle
\section{Introduction}
The ability to leverage GPUs to accelerate general-purpose workloads has resulted in unprecedented performance gains in various applications. Re-purposing shader cores primarily achieved this in GPUs, which accelerate arithmetic computations in rendering algorithms, to accelerate other non-rendering applications. However, re-purposing the cores was not an easy task: non-rendering applications needed to be rewritten as rendering algorithms to be able to use the shader cores. As only skilled graphics programmers were able to do this translation, researchers began to work on creating programming models that would allow any programmer to leverage GPU acceleration. This led to the creation of popular programming models such as CUDA\cite{cuda} and OpenCL\cite{open-cl}. They allowed programmers to write general-purpose code while abstracting the translations required to run it on the GPU. These programming models are responsible for cutting down execution times of machine learning applications from weeks to days and hours to seconds. 

While the shader cores were ideal for performing arithmetic computation on regular structures, they performed poorly on irregular applications such as tree traversal. In such cases, the user was often better off executing the application on the CPU. The introduction of Ray Tracing (RT) cores in recent GPUs created an opportunity to accelerate a different class of applications: irregular applications. {\em Ray casting} is a popular rendering algorithm where a ray is launched through a pixel in the image plane and the interaction of the ray with objects in the scene is used to determine the color of the pixel. {\em Ray tracing} expands on ray casting by also accounting for reflected and refracted rays (secondary rays) that are generated due to the ray-object interaction and trace the interaction of these secondary rays with the objects in the scene. The RT cores on GPUs were designed to accelerate the ray tracing the process by offloading certain sections of the RT pipeline to hardware, namely the Bounding Volume Hierarchy traversal and ray-bounding volume intersection tests~\cite{whitepaper} (Details in Section~\ref{sec:back-bvh}). 


Though the RT cores were designed to accelerate ray tracing applications, researchers are starting to look into exploiting these cores to perform general-purpose computations. Wald~\etal introduced the idea of using RT cores to identify the tetrahedral mesh to which a point belongs~\cite{wald19}. The authors re-formulate the problem as a ray tracing problem by treating the query point as the ray origin and the tetrahedral meshes as objects in the scene. Now, the point-in-tetrahedron problem reduces to launching a ray from the query point and identifying the closest intersecting mesh. Later works by Zellman~\etal, Evangelou~\etal, and Zhu~\etal looked at re-formulating the nearest neighbor search problem as a ray casting query ~\cite{force-directed-graph, Evangelou2021RadiusSearch, rtnn}. The Nearest Neighbor Search (NNS) problem is the task of identifying the closest points to a query point. Its variant, k-Nearest Neighbor Search (kNNS), restricts the task of finding the {\em k} closest neighbors. 

A major downside to RT-accelerated neighbor search is that the search space for nearest neighbors is constrained to a fixed-radius neighborhood due to the problem translation approach adopted by prior work (Details in Section~\ref{sec:fixed-rad-knn}). It is impossible to know the required radius to identify neighbors {\em a priori}, leading the approach to possibly find {\em less} than k neighbors. Prior work has suggested that this problem can be avoided by choosing a very large radius to ensure all neighbors are found \cite{rtnn}, but we show that this approach is highly inefficient---trying to find all of the neighbors of query points using the existing approaches obviates the benefits of hardware acceleration entirely (Details in Section~\ref{sec:eval}). 

This paper presents an efficient solution to the problem of RT core acceleration of {\em unbounded} kNNS---ensuring that all query points will successfully find all $k$ neighbors. We adopt an iterative solution: we start with a smaller search radius and keep track of points that have found their {\em k} nearest neighbors in each iteration. In the subsequent iterations, we incrementally increase the radius and {\em only} query the points that have not found their neighbors. While this approach seems like it should be significantly slower than choosing a single radius, we show that it is not. Since the number of points being queried decreases as the search radius increases, we find that our approach is significantly {\em faster} than choosing an arbitrarily large radius, and can even be faster than prior fixed-radius approaches even for smaller radii. 

To summarize, the contributions of our paper are as follows:
\begin{itemize}[topsep=0pt]
    \item This paper introduces TrueKNN, the first RT-accelerated neighbor search algorithm that is not constrained to a fixed-radius.

    \item We show that TrueKNN outperforms fixed-radius, non-iterative approaches by large margins. This iterative solution gradually grows
the neighbor search space while pruning query points
that have already found their neighbors, leading to
significantly fewer ray-object intersection tests

    \item We further show that, unlike prior approaches that require {\em a priori} selection of a query radius, we can adapt TrueKNN to find the appropriate radius dynamically, outperforming prior approaches even when those approaches select their query radius {\em a posteriori}.

\end{itemize}

\section{Background}
In this section, we introduce the k-Nearest Neighbor Search (kNNS) problem and explain how prior works have translated kNNS to a ray casting problem.
\subsection{k-Nearest Neighbor Search}
The Nearest Neighbor Search (NNS) problem was first introduced by Fix\etal~\cite{KNN-Fix1989DiscriminatoryA} and expanded by Cover\etal~\cite{knn-newer}. It is defined as follows:

\begin{definition}
For a dataset $D$ and query point $q \in D$, find the set of {\em k} nearest points to $q$.
\end{definition}

The nearest neighbors are typically identified using a distance metric, with Euclidean distance being the most popular choice. kNNS is primarily used as a subroutine in k-Nearest Neighbor classification and regression algorithms. The basic idea behind these algorithms is that a property of a query point can be determined by observing its nearest neighbors. For example, a query point can be classified into the same class as a majority of its neighbors in classification problems. Similarly, the properties of a query point can be averaged using its neighbors in regression problems. 

kNN classifier and regression models are widely used in point cloud applications to compute surface normals~\cite{point-cloud}, recommendation systems to assign recommendations based on similar users~\cite{knn-recomm-ADENIYI201690}, healthcare to classify patients~\cite{knn-health} and pattern recognition, to name a few. 

\subsection{Ray Tracing Hardware} \label{sec:back-rt}
As a part of NVIDIA's Turing architecture, each Streaming Multiprocessor (SM) has an RT core to accelerate BVH traversal and perform ray-triangle intersection tests, allowing the SM to perform other computations in the meantime. Both the SM and RT core share the same memory, allowing us to use both units in parallel. We direct the reader to \cite{whitepaper} for more details.

\subsubsection{Ray Casting}
The RT cores are able to accelerate the ray casting process by reducing the {\em number} of intersection tests performed, and this is facilitated by the Bounding Volume Hierarchy (BVH) structure used for object representation. The ray-casting process involves launching rays from a source through each pixel in the image plane, recording their interactions with objects in the scene, and using that information to determine the color of the pixels. It would seem that {\em each ray} would have to be tested for intersection against {\em each object} to check whether an intersection could affect the color of the pixel. However, this would be very inefficient since rays may not intersect a large subset of objects and intersection tests are computationally very expensive. 

\subsubsection{Bounding Volume Hierarchy} \label{sec:back-bvh}
Bounding Volume Hierarchies are an acceleration structure used to reduce the number of required ray-object intersection tests. The general idea is as follows: if we group objects that are spatially close to each other, we can test for intersection against {\em groups} of objects rather than {\em individual} objects, reducing the number of intersection tests performed. The idea is to enclose objects in bounding volumes and recursively enclose these volumes in larger bounding volumes until we create a volume that is large enough to enclose the entire scene. These bounding volumes are represented hierarchically using a tree structure called the Bounding Volume Hierarchy (BVH). Each node in the tree represents a bounding volume that encloses all its descendent nodes. The most commonly used bounding volume is an Axis-Aligned Bounding Box (AABB).

\begin{figure*}[ht]
    \centering
    \includegraphics[width=0.85\textwidth]{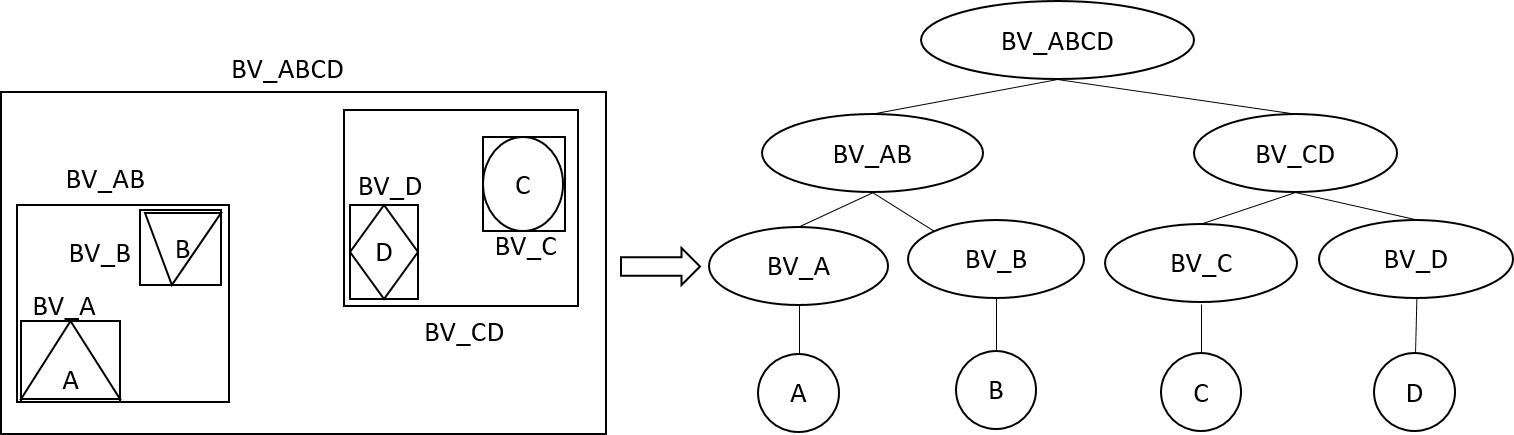}
    \vspace{-0.5em}
    \caption{Bounding Volume Hierarchy Construction. The scene with objects enclosed in bounding volumes is shown on the left. On the right, the BVH corresponding to the objects and bounding volumes is shown.}
    \label{fig:bvh}
    \vspace{-0.8em}
\end{figure*}
Figure~\ref{fig:bvh} shows how objects in a scene are used to construct the BVH. On the left, each object A, B, C and D are enclosed in their corresponding bounding volumes BV\_A, BV\_B, BV\_C and BV\_D. As A and B are spatially close together, they are combined into a larger bounding volume BV\_AB that encloses the bounding volumes of A and B. BV\_CD is constructed similarly and BV\_ABCD encloses both BV\_AB and BV\_CD to capture the entire scene. The hierarchical relationship between the bounding volumes is captured on the right. For example, BV\_AB encloses BV\_A and BV\_B, and this relationship is captured in the BVH with BV\_AB as the parent node and BV\_A and BV\_B as the children node.

It is possible to reduce the number of ray-object intersection tests by performing ray-{\em AABB} intersection tests and only performing the ray-{\em object} intersection test if the previous test succeeds. For example, in Fig~\ref{fig:bvh}, if the ray does not intersect BV\_AB, it is {\em guaranteed} to not intersect BV\_A and BV\_B. Since A and B are contained in these bounding volumes, we do not have to perform the corresponding ray-{\em object} intersection tests. This approach allows us to prune large parts of the search space since if the ray does not intersect the AABB, it is guaranteed to not intersect any of the objects or other bounding volumes contained in the AABB.

\subsubsection{Optix API} \label{sec:optix-api}
\begin{figure}
    \centering
    \includegraphics[width=\linewidth]{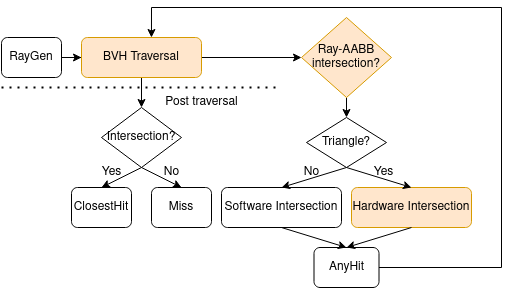}
    \caption{Ray Tracing Pipeline in Optix API}
    \label{fig:optix-api}
\end{figure}
The Optix API handles offloading Bounding Volume Hierarchy build and traversal to the RT core, while allowing the user to write custom CUDA kernels that use the GPU's shader cores. 

Fig~\ref{fig:optix-api} shows the working of the Optix API. The {\em RayGen} program is responsible for creating rays with the specified origin ($\Vec{o}$), direction ($\Vec{d}$), and length ($t$)
\[\Vec{r} = \Vec{o} + t\Vec{d}, t \in [t_{min}, t_{max}]\]
The ray traverses the BVH and checks for intersection with the Axis-aligned Bounding Boxes (AABBs) in hardware (Highlighted in yellow in Fig~\ref{fig:optix-api}). The ray-{\em object} intersection test can happen in software or hardware, depending on the object. If the object is a triangle, the test happens in hardware. If not, the user provides a custom intersection test written as a software CUDA {\em Intersection} program. Optix allows us to perform the ray-AABB intersection test in hardware and the ray-object intersection test in software/hardware. 

The user can evoke the {\em AnyHit} program to record intersections, decide whether to continue or terminate the BVH traversal for a particular ray and launch subsequent rays.
After the BVH traversal completes entirely, the user can specify a {\em ClosestHit} to record the closest intersected object to the ray. The user can also specify what to do in case of no intersections using the {\em Miss} program.

\subsection{RT-accelerated k-Nearest Neighbor Search (RT-kNNS)} \label{sec:fixed-rad-knn}
We use the reduction proposed by Zellman\etal to translate fixed-radius nearest neighbor queries to ray tracing queries~\cite{force-directed-graph}. The translation relies on a key observation: to find neighbors within a radius {\em r} of point {\em p}, we can expand a sphere of radius {\em r} around {\em all} points and check how many spheres contain the point {\em p}. The centers of the $k$-closest spheres are the neighboring points of {\em p}. 

\begin{algorithm2e}
\caption{{\tt RT-kNNS}}\label{alg:rt-knn}
\SetKwInOut{Input}{Input}
\SetKwInOut{Output}{Output}
\Input{Dataset {\em D}, radius {\em r}}
\Output{Neighbors within radius {\em r}}
$spheres \gets createSpheres(D, r)$ \\
$boundedSpheres \gets createAABB(spheres)$ \\
$constructBVH(boundedSpheres)$ \\
\For{Point $p \in D$}
{
    $ray \gets RayGen(\Vec{p}, \Vec{d}, 0, ${\tt FLOAT\_MIN}) \\
    \While{$traverseBVH(ray)$} 
    {
        \If{$Intersect(ray, AABB)$} 
        {
            \If{$Intersect(ray, sphere)$} 
            {
                $p.neighbor \gets p.neighbor \bigcup ray.center$
            }
        }
    }
}
\end{algorithm2e}

RT-accelerated kNNS is outlined in Algorithm~\ref{alg:rt-knn} and is based on the Optix API pipeline from Section~\ref{sec:optix-api}. In Line 1, we create spheres for all points with $p \in D$ ({\em D} is the input dataset). The centers of the sphere are the points $p$ and the radius is supplied by the user. In Line 2, we specify a {\em BoundingBox} program to create Axis-Aligned Bounding Boxes (AABBs) to enclose the spheres. We proceed to construct a Bounding Volume Hierarchy (BVH) by recursively combining bounding volumes (See Section~\ref{sec:back-bvh}). 

With the BVH constructed, we can begin to generate rays that traverse the BVH and test for ray-AABB and ray-object intersections. We create rays for all points $p \in D$ in Line 5. The {\em RayGen} program accepts origin, direction, and ray interval ($t_{min}, t_{max}$) as inputs. We set point $\Vec{p}$ as center, (0,0,1) as the direction ($\Vec{d}$) and $t_{max}$ as {\tt FLOAT\_MIN}. Since a ray of infinitesimal length is sufficient to intersect neighbors, $t_{max}$ can be set to a very small value. In Lines 6-12, each ray traverses the BVH and performs intersection tests against AABBs. If the ray intersects the AABB, ray-object intersection tests are performed for {\em each} object contained in the intersected AABB. If the ray intersects the sphere, we add the sphere's center to the list of point {\em p}'s neighbors.

\section{Design} \label{sec:design}
In this section, we describe how we use the fixed-radius k-Nearest Neighbor Search (kNNS) reduction from Section~\ref{sec:fixed-rad-knn} to implement our {\em unbounded} RT-accelerated kNNS algorithm, TrueKNN. TrueKNN incrementally increases the search space till we find each point's $k$ nearest neighbors.

\subsection{TrueKNN Overview} \label{sec:design-trueknn}
We define TrueKNN as the task of finding the $k$ nearest neighbors of all points in a dataset. TrueKNN differs from fixed-radius kNNS in that we do not constrain our neighbor search space to a particular neighborhood. Section~\ref{sec:fixed-rad-knn} shows how to reduce fixed-radius nearest neighbor queries to ray tracing queries by expanding spheres with a user-specified radius around all points. In TrueKNN, the user does not specify the radius.

Since the user does not have to specify the radius, we need to figure out a radius such that each point in the input dataset can find all its neighbors. A solution proposed by Zhu is to use an arbitrarily large radius for the sphere expansion phase~\cite{rtnn}. This would guarantee that each point finds its $k$ nearest neighbors. However, choosing this appropriately-large radius is difficult {\em a priori}: without performing kNNS, to begin with, we do not know how large a radius will guarantee that all points will find their $k$ nearest neighbors. The only radius that is guaranteed to work {\em a prior} is one that encompasses {\em all} the points. But in such a scenario, the volumes around each point are guaranteed to overlap as the bounding boxes become larger when the search radius is large. As the bounding volume hierarchy can no longer separate AABBs, the algorithm is definitive $O(n^2)$.

So {\em a priori} radius setting cannot work. Instead, our approach is to incrementally identify the radius needed to identify all of the neighbors. We propose a multi-round fixed-radius kNNS approach that implements TrueKNN. We begin by performing fixed-radius kNNS with a small radius and determine the starting radius for the first round of our neighbor search by selecting the minimum distance between points in a subset of the input dataset (Section~\ref{sec:random_sample}). Because this radius is small, the search completes quickly, but many points may not find their k neighbors. In each subsequent round, we increment the radius and re-run fixed-radius kNNS on any points that have not yet found their $k$ neighbors until all points find their $k$ nearest neighbors (Section~\ref{sec:multi-round}).

\subsection{Determining start radius} \label{sec:random_sample}
In Algorithm~\ref{alg:rt-knn}, the user specifies the radius for sphere expansion. However, in the case of TrueKNN, there is no user input and we need to determine an appropriate start radius to begin the neighbor search. The selection of a start radius is crucial to performance. If the radius is too small, we will have many rounds where no points find any neighbors and we will pay for the cost of a context switch between device and host, BVH refit, ray launch, and intersection tests. On the other hand, if the radius is too large\footnote{ We define the notion of small and large radius in comparison to the maximum distance between a point and any of its $k$ nearest neighbors in the input dataset.}, the number of intersection tests performed will increase by several orders of magnitude, leading to poor performance.

\begin{algorithm2e}
\caption{{\tt RandomSample}}\label{alg:random-sample}
\SetKwInOut{Input}{Input}
\SetKwInOut{Output}{Output}
\Input{Dataset {\em D}}
\Output{Radius {\em start\_radius}}

random\_sample $\gets sample$({\em D}, 100) \\
neighbors, distances $\gets kNearestNeighbors$({\em D}, $k$=4) \\
start\_radius $\gets min$(distances)
\end{algorithm2e}

It was evident that we needed to have some idea about the input dataset to select a good start radius. Without this information, it would be impossible to know whether our chosen radius is too small or large. To incorporate dataset information, we propose a random sampling approach (Algorithm~\ref{alg:random-sample}) to find the minimum distance between neighbors of a subset of the input dataset. In Line 1, we choose 100 random points from the input dataset. We then use Python scikit-learn's built-in, ball-tree based $k$ nearest neighbors algorithm to find the 4 nearest neighbors of these randomly sampled points. We empirically chose $k=4$ as it worked well in our experimental evaluation and had negligible execution time (5 to 8 ms). We note that increasing $k$ and/or sample size {\em could} result in a better start radius. As shown in Line 3, we then find the minimum distance between a point and its neighbors and set that as our staring radius. 

We experimented with different starting radii and found that the cost of choosing a larger radius was much higher than starting off with a smaller radius. When starting off with a very small radius, we observe that some points find few to no neighbors in the initial rounds. However, we found that the time taken by these initial rounds was insignificant compared to the total execution time. On the other hand, starting off with a larger radius leads to fewer rounds but more intersection tests, justifying our decision to use the minimum distance as the start radius. We show that our random sampling approach produces useful starting radii in Section~\ref{sec:eval-start-radius}.

\subsection{Multi-round kNNS} \label{sec:multi-round} 
Now that we have a starting radius, Algorithm~\ref{alg:true_knn} describes our multi-round approach for TrueKNN.
\begin{algorithm2e}
\caption{{\tt TrueKNN}}\label{alg:true_knn} 
$radius \gets$ {\tt RandomSample}(D) \\
\While{$D \ne \{\emptyset\}$} 
{
    
    $neigh, dist \gets$ RT-kNNS$(D, radius)$ \\
    \For{$p \in D$}
    {
         \If{$|p.neigh| == k$} 
            {
                $D -= \{p\}$
            } 
            
     }  
     \If{$D \ne \{\emptyset\}$}
     {
        $radius \gets radius * 2$\\
        {\tt REFIT\_BVH}$(D, radius) $\\
     }
}

\end{algorithm2e}

We use the random sampling approach outlined in Section~\ref{sec:random_sample} to determine the radius for sphere expansion in the first round of fixed-radius RT-kNNS. We use the minimum neighbor distance returned by {\tt RandomSample} as our start radius ($radius$) in Line 1. In Line 3, we call RT-kNNS with our dataset (D) and radius ({\em radius}) to find all points within a fixed neighborhood of {\em p}. We then check if the previously chosen radius was sufficient for all points to find $k$ neighbors in Line 5. In Line 6, we remove all the points that have found their $k$ neighbors from our dataset so that the next iteration will only consider points that have not found their neighbors yet. For points that have not found all $k$ neighbors, we expand the neighbor search space by incrementing the radius of spheres and re-fitting the bounding boxes around the spheres for the BVH, as shown in Lines 10 and 11. We continue this process till $D = \{\emptyset\}$ and all points have found their $k$ nearest neighbors.

\subsection{TrueKNN Discussion}
It may seem curious that this algorithm is more efficient than prior fixed-radius approaches---after all, each iteration of TrueKNN performs an {\em entire fixed-radius} nearest neighbor search. The key is in the interplay between radius size, BVH-based search speed, and sorting time. When the radius is small, the spheres around the points are small and well-separated. As a result, the BVH is extremely effective at accelerating ray tracing, and query points very quickly identify nearby points. By starting with a small radius, this first iteration is {\em faster} than ``normal'' fixed-radius kNNS, with the trade-off that many points are unable to find their needed k neighbors.

In subsequent iterations, the radius is increased, but because {\em some} points in earlier iterations have already found their needed neighbors, those points do not need to be re-queried. So although the ray tracing process is slower, there are fewer query points, and the overall search is faster. In the final iterations, when the radius is quite large and BVH acceleration is essentially useless, there are only a few query points remaining (the outliers).

As mentioned earlier, setting a large starting radius for fixed-radius kNNS leads to inefficient filtering, and a $O(n^2)$ runtime. By incrementally increasing the radius, TrueKNN avoids this problem---while outlier points have to do an $O(n)$ search to find their candidate neighbors, most points are resolved with smaller radii and require closer to a $O(\log n)$ search.

There is another beneficial effect of TrueKNN's iterative approach. With a large starting radius (e.g., enough to find at least k neighbors for 99\% of the points), {\em many} points will find {\em too many} candidate neighbors in their initial search, and hence will waste time sorting those candidates to find the k closest. By resolving those points with smaller radii, TrueKNN also reduces sorting time. Indeed, as we show in Section~\ref{sec:eval-99th}, TrueKNN's benefits are not only because outliers force large query radii for fixed-radius search. Even in settings where we do not attempt to resolve outlier points, TrueKNN outperforms fixed-radius kNNS.


\section{Implementation Details}
In this section, we discuss the implementation of TrueKNN using the Optix API. We used the Optix Wrapper Library (OWL)~\cite{owl}, which is built on top of Optix 7, to implement TrueKNN and our baseline. OWL allows the user to define custom shader programs to test for the ray-object intersection. Though one would typically use the {\em AnyHit} program (Section~\ref{sec:optix-api}) to collect multiple hits, we implemented the TrueKNN logic in the {\em Intersection} program to avoid incurring overhead costs associated with calling the {\em AnyHit} program. In fact, we disable both the {\em AnyHit} and {\em ClosestHit} program invocations to avoid performance penalties.

Since TrueKNN (Algorithm~\ref{alg:true_knn}) increases the radius of spheres in every iteration to expand the search space, the BVH corresponding to the objects also needs to change every iteration. One way to handle this is to re-{\em build} the BVH in every iteration. However, Optix\footnote{We use OWL and Optix interchangeably as every feature in OWL is also present in Optix} provides the option of BVH {\em re-fit}. This allows us to {\em re-fit} the bounding volumes to accommodate the expanded objects without having to explicitly re-build the BVH. We found that re-fitting was between 10-25\% faster than re-building.

\section{Evaluation} \label{sec:eval}
In this section, we evaluate TrueKNN's performance by analyzing the effect of varying parameters such as dataset size and $k$. We also look at how outliers in the dataset affect both TrueKNN and the baseline.

\subsection{Datasets}
We used a mixture of 2D and 3D real-world datasets (3DRoad, Porto, KITTI, and 3DIono) and a 3D synthetic dataset (UniformDist) to evaluate TrueKNN.
\begin{description}
    \item[3DRoad] 
    The 3DRoad dataset captures the road network of North Jutland, Denmark~\cite{3droad}. The dataset consists of {\em 430K} points. We use this as a 2D dataset, using only the latitude and longitude parameters.

    \item[Porto]
    The Taxi Service Trajectory - Prediction Challenge 2015 dataset captures vehicle movement trajectory data of 442 taxis in the city of Porto, Portugal~\cite{porto}. The dataset has just over {\em 81M} points and we use Porto as a 2D dataset, using only the latitude and longitude parameters.

    \item[KITTI]
    The KITTI vision benchmark dataset captures data from the movement of an autonomous vehicle around the city of Karlsruhe, Germany~\cite{kitti-Geiger2013IJRR}. The dataset has just over {\em 1M} points and we use KITTI as a 3D dataset.

    \item[3DIono]
    The 3D Ionosphere dataset captures the behavior of electrons in the ionosphere~\cite{3diono}. The dataset has just over {\em 1M} points and we use it as a 3D dataset.

    \item[UniformDist]
    We create a synthetic 3D dataset of {\em 1M} points that is uniformly distributed on [0,1] to study the impact of outliers on our algorithm.
    
\end{description}

\subsection{Experimental Setup}
 We ran our experiments on an NVIDIA GeForce RTX 2060 GPU with 6 GB device memory, CUDA version 10.1, and Optix 7.1. As Optix is a graphics rendering API, it accepts only 3D input data. As a workaround,  we set the z-dimension to 0 for 2D datasets.

\subsubsection{Baseline} \label{sec:baseline}
We use RT-kNNS (Algorithm~\ref{alg:rt-knn}) as our baseline by setting the radius for sphere expansion as the maximum distance ($maxDist$) between a point and any of its $k$ nearest neighbors. This way, the baseline is guaranteed to find all $k$ nearest neighbors of each point in the dataset. We chose this baseline as prior work has shown that RT-accelerated neighbor search is consistently faster than other GPU-based implementations~\cite{rtnn,Evangelou2021RadiusSearch,wald19}. We also note that our baseline represents the {\em best case} scenario since our neighbor search is constrained exactly to a $maxDist$-neighborhood. In practice, the user would probably select some arbitrary $d$-neighborhood, where $d >> maxDist$.  


\subsection{Performance Evaluation}
We compare the performance of TrueKNN against RT-kNNS as described in Section~\ref{sec:baseline}. We study the performance impact of dataset size by varying the dataset size between {\em 100K} and {\em 1M}. For each dataset size, we chose $k = 5$ and $k = \sqrt{Dataset Size}$ to study the impact of varying $k$. We chose $k = \sqrt{DatasetSize}$ as it is the commonly used $k$ value for KNN classifier and regression models~\cite{k-value-NADKARNI2016187}. We only vary dataset size up to {\em 1M} since we run out of memory to store neighbors in our GPU when $k = \sqrt{Dataset Size}$. We also perform an experiment where we modify the kNNS problem to one of finding $99^{th}$ percentile neighbors and evaluate our performance against the baseline to understand the effect of outliers in the dataset.

For different dataset sizes ({\em d}), we always used the first {\em d} points in our experiments and averaged results over 5 runs. The reported execution time for all our experiments includes BVH build and refit times, context switching overheads and neighbor search time. We assume that the input is already on the device and do not include data transfer time.

\subsubsection{Impact of Dataset size} \label{sec:eval-data-size}
\begin{figure}[ht]
    \centering
    \includegraphics[width=0.45\textwidth]{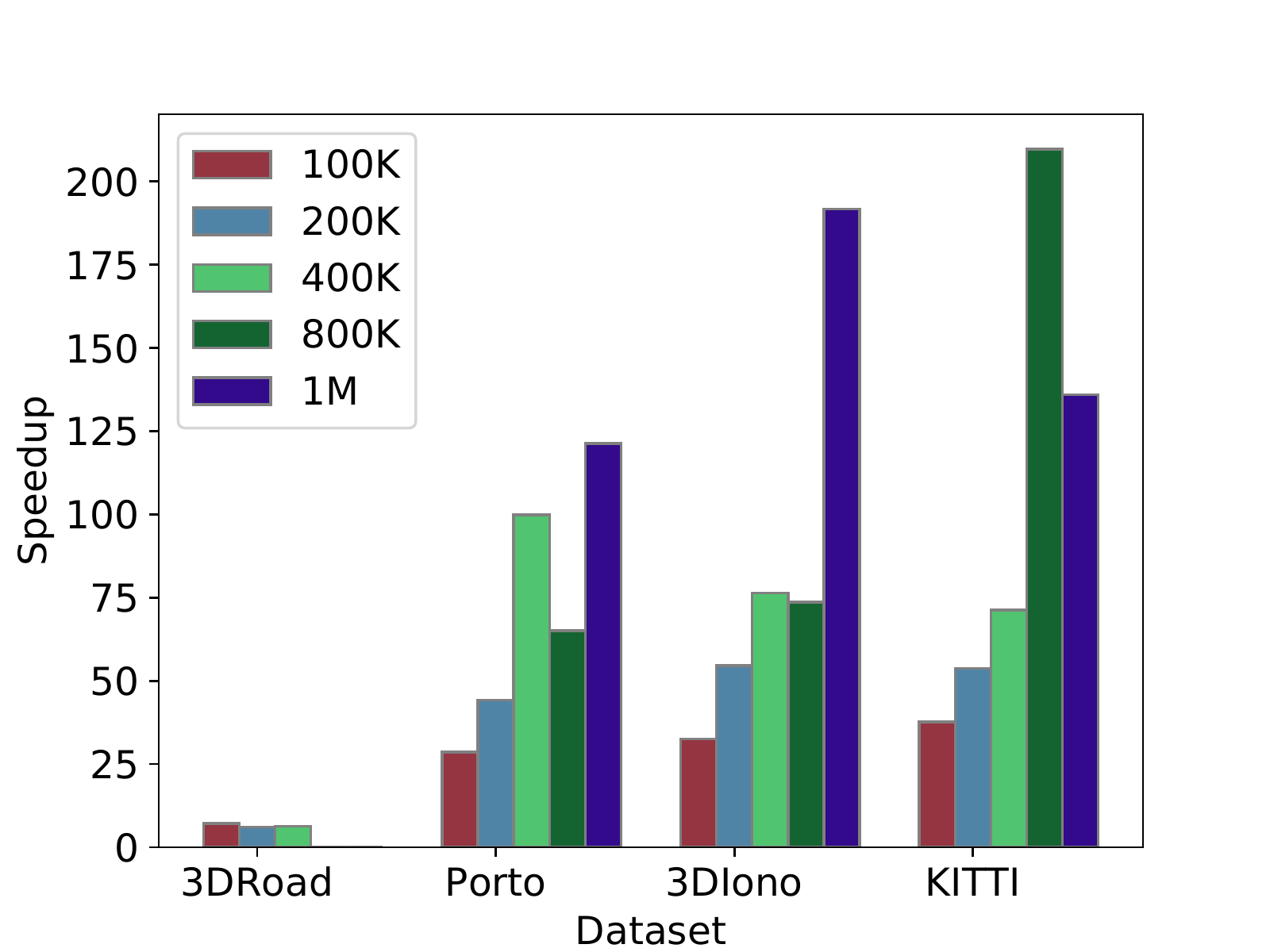}
    \vspace{-0.5em}
    \caption{TrueKNN's speedup compared to baseline while varying dataset size}
    \label{fig:dataset-size}
    \vspace{-0.8em}
\end{figure}

We study the effect of varying the number of query points for a fixed number of neighbors ($k = \sqrt{DatasetSize}$). From Fig~\ref{fig:dataset-size}, we see that TrueKNN outperforms the baseline on all datasets\footnote{ 3DRoad does not have speedup bars for {\em 800K} and {\em 1M} as it only has {\em 400K} points in total}. Table~\ref{table:exec-time} also shows the raw execution time for all datasets. A major advantage of TrueKNN is that points that have found their $k$ nearest neighbors in previous rounds are no longer queried in subsequent rounds with larger search radii. As the search radius increases, the size of the Axis-Aligned Bounding Boxes (AABBs) enclosing the spheres also increases. This leads to an increase in ray-AABB intersection tests, which, in turn, leads to an increase in ray-object intersection tests. Since TrueKNN can correctly identify the $k$ nearest neighbors, any additional intersection tests (compared to TrueKNN) performed by the baseline are unnecessary. As the number of query points increases, we see that the number of these unnecessary intersection tests also increases.

\begin{table*}[h]
    \begin{tabular}{lcccccccccccl} 
       \toprule
       Dataset size  & 
       \multicolumn{2}{c}{3DRoad} & \multicolumn{2}{c}{Porto} & \multicolumn{2}{c}{3DIono} & \multicolumn{2}{c}{KITTI}\\

       & TrueKNN & Baseline  & TrueKNN & Baseline
       & TrueKNN & Baseline  & TrueKNN & Baseline\\
       \midrule 
       100K & 6.28 & 44.71 & 25.27 & 722.93  &  9.48 & 308.16 & 13.8 & 520.01 \\

       200K & 23.97 & 146.83 & 89.72 & 3965.7 &  38.16 & 2086.13 & 51.6 & 2769.96\\

       400K & 120.41 & 753.81 & 456.78 & 45639.7 & 149.6 & 11433.5 & 175.38 & 12503.4\\
       
       800K &  - & - & 1128.67 & 73423.5 &776.96 & 57184.7 &641.96 & 776562.4\\
       
       1M &  - & - & 1052.95 & 127720.1 &973.3 & 1862527.9 & 892.21 & 121256.2\\
       \bottomrule
   \end{tabular}
   \caption{Execution time (in seconds) for TrueKNN and baseline for all datasets}
\label{table:exec-time}
\vspace{-1em}
\end{table*}

Since the ray-AABB intersection tests are performed in hardware, we do not have any information on the number of tests performed. However, since the ray-{\em object} intersection tests are performed in software, we can compare the number of intersection tests performed by TrueKNN and the baseline. Table~\ref{table:intersections} shows the number of ray-sphere intersection tests performed (in billions) on the Porto dataset. We notice that the baseline performs 9x the tests performed by TrueKNN for {\em 100K} points and this increases to 32.1x for {\em 1M} points. As TrueKNN incrementally increases the search radius and reduces the number of query points launched every round, we perform far fewer intersection tests compared to the baseline, leading to significant performance improvements. 

Though this gives us an idea of where our speedup comes from, speedup does not always increase proportionally to the decrease in the number of intersection tests. For example, the baseline performs 16.8x and 24.7x more intersection tests compared to TrueKNN for {\em 400K} and {\em 800K} points, respectively. However, our speedup falls from 99.9x to 65x. There are possibly other RT Core-related aspects (ray-{\em AABB} intersection tests, BVH traversal) that contribute to this\footnote{At the time of writing this paper, there were no tools available to profile the RT Cores}.
\begin{table}[h!]
\centering
    \begin{tabular}{lcccl} 
    \toprule
       Dataset size  & TrueKNN & Baseline \\
        \midrule
       100K & 1.09 & 9.9 \\

       200K & 2.85 & 39.9 \\

       400K & 7.16 & 189.1 \\

       800K & 25.86 & 639.31 \\

       1M & 31.12 & 999.19 \\
       \bottomrule 
   \end{tabular}
   \caption{Number of ray-object intersection tests performed (in billions) for the Porto dataset}
\label{table:intersections}
\vspace{-0.8em}
\end{table}

\paragraph{Comparison with RTNN}
RTNN was proposed by Zhu to optimize nearest neighbor search using RT cores~\cite{rtnn}. RTNN proposes two main optimizations (1) query reordering to improve ray coherence (2) query partitioning to minimize intersection tests. To demonstrate the effectiveness of our approach, we compare an unoptimized TrueKNN (does not use query sorting or partitioning optimizations) against a fully optimized RTNN on the Porto dataset. We found that TrueKNN was between 1.5x to 8x faster than RTNN, showing that our ability to perform fewer computations results in huge performance improvements.

\paragraph{Comparison with non-RT Baseline}
We evaluate TrueKNN's performance against cuML's kNNS implementation, which is a purely CUDA-based implementation~\cite{cuml-raschka2020machine}. As cuML's kNNS ran out of memory when we set $k = \sqrt{DatasetSize}$, where $Datasetsize > 200K$, we chose $k = 5$ for our experiment. 

\begin{figure}[ht]
    \centering
    \includegraphics[width=0.45\textwidth]{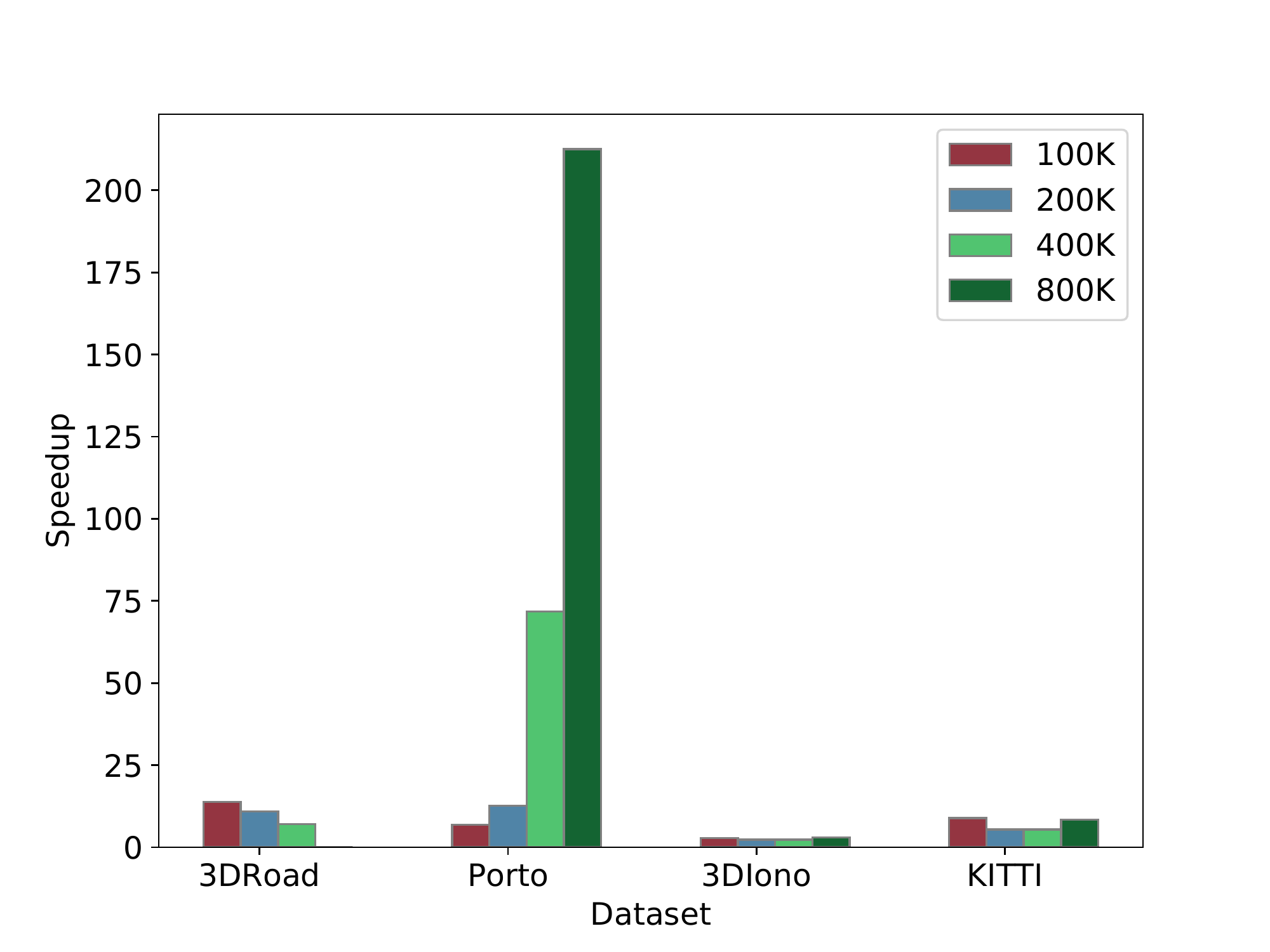}
    \vspace{-0.5em}
    \caption{TrueKNN's speedup compared to cuML's kNNS implementation on varying dataset size}
    \label{fig:cuml-dataset-size}
    \vspace{-0.8em}
\end{figure}
From Fig~\ref{fig:cuml-dataset-size}, we see that TrueKNN outperforms cuML's kNNS across all the different datasets and dataset sizes. We attribute this to TrueKNN's ability to leverage hardware acceleration while reducing the computation footprint. In general, we found that TrueKNN's speedup increased on increasing the dataset size. 

\subsubsection{Impact of $k$} \label{sec:eval-vary-k}
\begin{figure}[ht]
    \centering
    \includegraphics[width=0.45\textwidth]{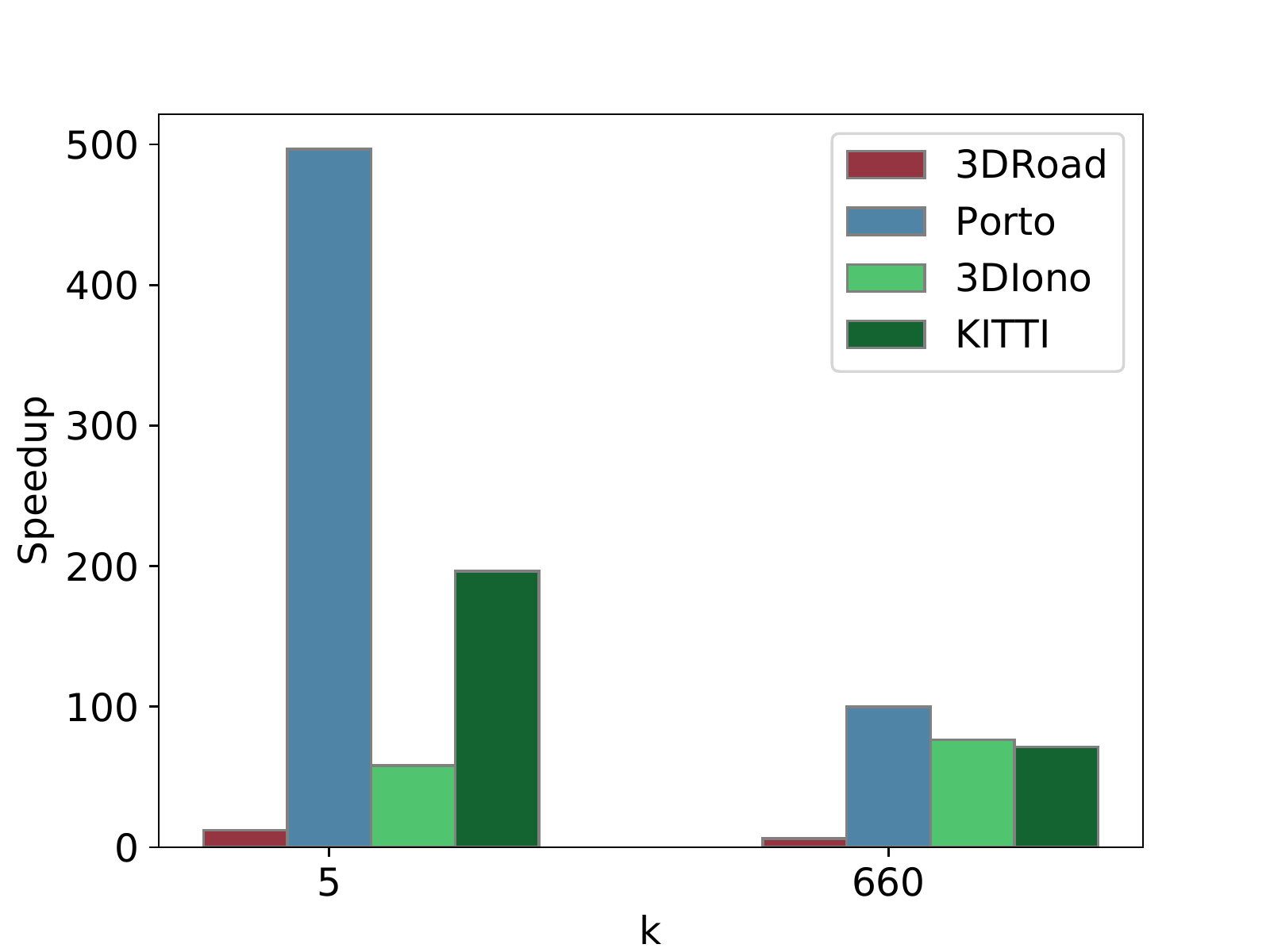}
    \vspace{-0.5em}
    \caption{Impact of k}
    \label{fig:vary-k}
    \vspace{-0.8em}
\end{figure}

We set the $k$ parameter as 5 and \\$\sqrt{DatasetSize}$ for {\em 400K} points from the 3DRoad, Porto, 3DIono, and KITTI datasets. In both cases, we consistently outperform the baseline, as shown in Fig~\ref{fig:vary-k}. We also notice that the extent of our speedup is larger when $k=5$ compared to $k=660$. We mainly attribute this to the overhead of sorting and maintaining the list of $k$ nearest neighbors over multiple rounds in the case of large $k$ values, compared to the one-time cost incurred by the baseline. 

Additionally, we notice that as $k$ increases, the number of ray-object intersection tests also increases as we need to identify more neighbors. For example, on the Porto dataset, the baseline performs 839x the intersection tests performed by TrueKNN when $k=5$ but this number decreases to 17x when $k=660$.
However, we are still over 70x faster than the baseline for most datasets when $k=660$.

\subsection{Performance Analysis}
In this section, we justify our design decisions by analyzing the run time of the various rounds in our TrueKNN approach and showing that our choice of start radius yields good results.

\subsubsection{Runtime Analysis} \label{sec:eval-runtime-analysis}
\begin{figure}
     \centering
     \begin{subfigure}[b]{0.45\textwidth}
         \centering
         \includegraphics[width=\textwidth]{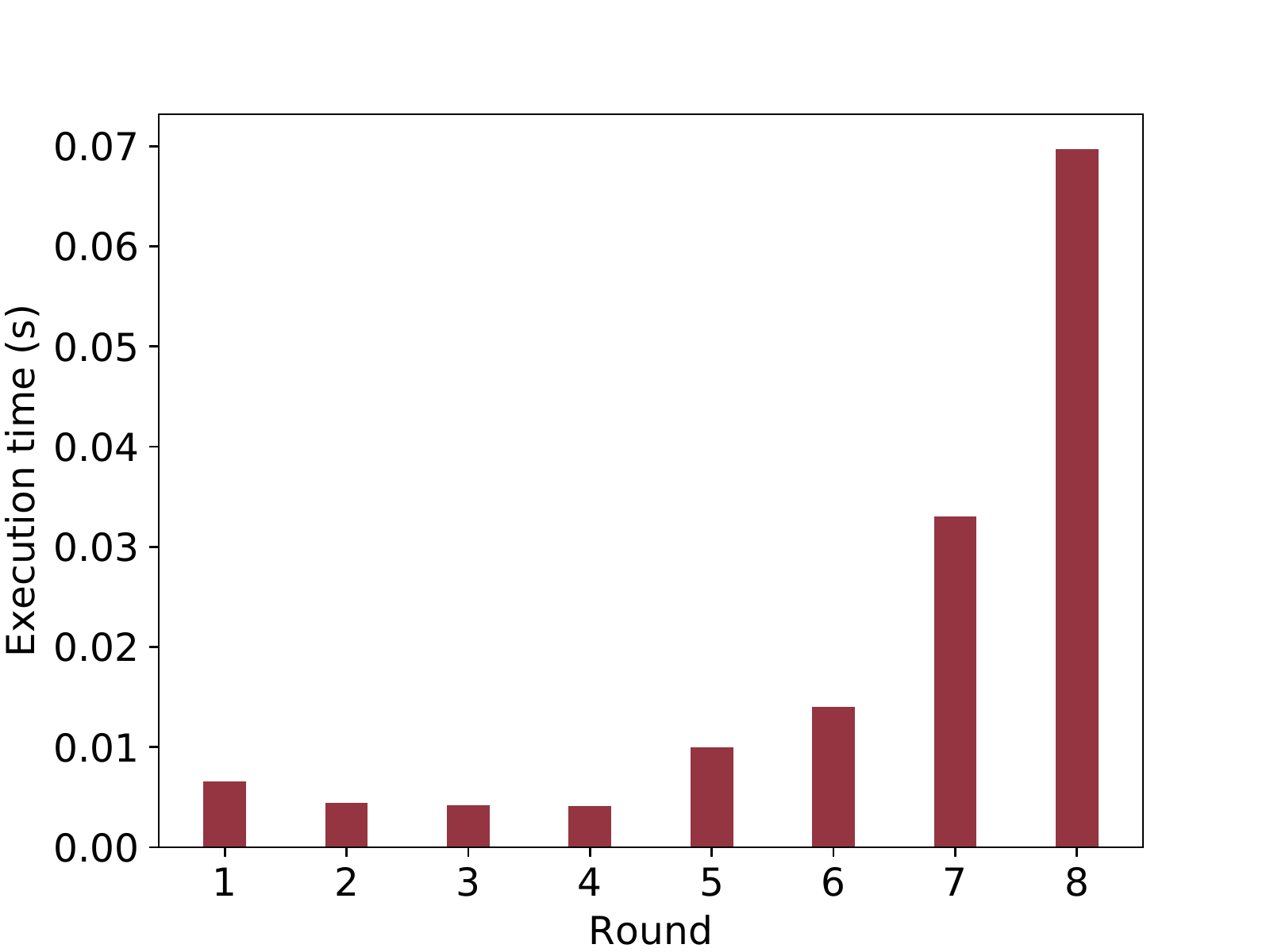}
         \caption{Time taken by each round}
         \label{fig:time-round}
     \end{subfigure}
     \hfill
     \begin{subfigure}[b]{0.45\textwidth}
         \centering
         \includegraphics[width=\textwidth]{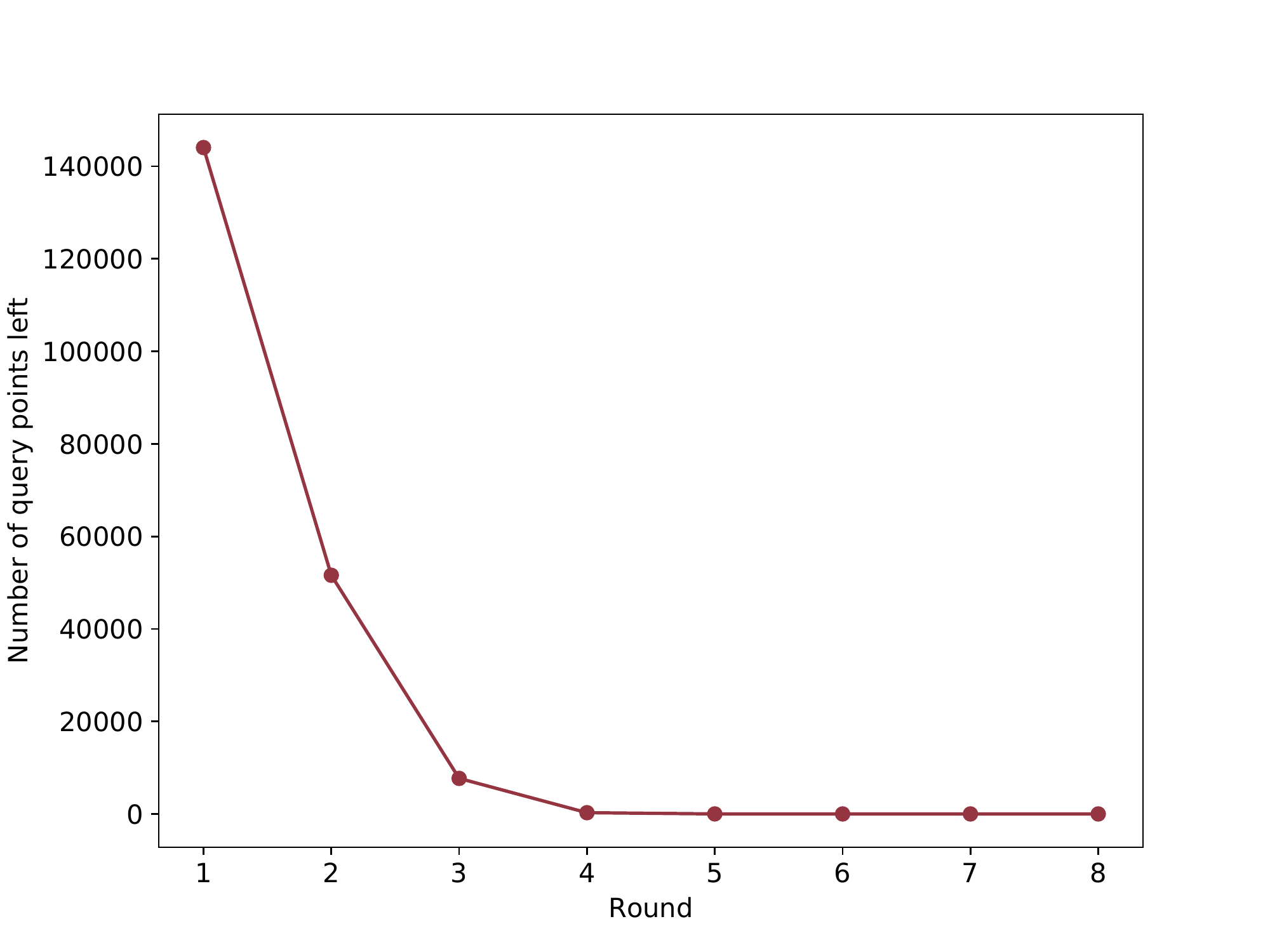}
         \caption{Number of points left after each round}
         \label{fig:num_points-round}
     \end{subfigure}
     \hfill
        \caption{Breakdown of 3DRoad execution time} 
        \label{fig:stats-breakdown}
        \vspace{-1em}
\end{figure}
To understand why TrueKNN is able to largely outperform the baseline, we evaluated TrueKNN on the first 400K points of the 3DRoad dataset, with a start radius of 0.001 and $k=5$. Fig~\ref{fig:time-round} shows the time taken for each round of TrueKNN and Fig~\ref{fig:num_points-round} shows the number of query points that {\em have not} found their $k$ nearest neighbors after each round. 

From Fig~\ref{fig:time-round}, it is evident that rounds 7 and 8 take significantly more time compared to the other rounds. This is interesting, since these rounds query far fewer points compared to earlier rounds, as most points have already found their $k$ neighbors. Indeed, from Fig~\ref{fig:num_points-round}, we see that only 3 query points remain in these last few rounds. Though one would expect that fewer query points would lead to faster execution times, we find that it is not true because the search radius is much larger in the latter rounds.

As the search radius increases in latter rounds, even if we have only 3 query points, the BVH traversal will result in a large number of intersection tests as the size of the bounding volumes enclosing the spheres also increases to accommodate the larger sphere. As we double the radius in each round, we see a huge uptick in the number of intersection tests. This shows us why TrueKNN is able to outperform the baseline: we expand {\em fewer} points to larger radii in latter rounds, allowing us to drastically reduce the number of intersection tests performed.

\subsubsection{Start Radius}\label{sec:eval-start-radius}
\begin{figure}[ht]
    \centering
    \includegraphics[width=0.45\textwidth]{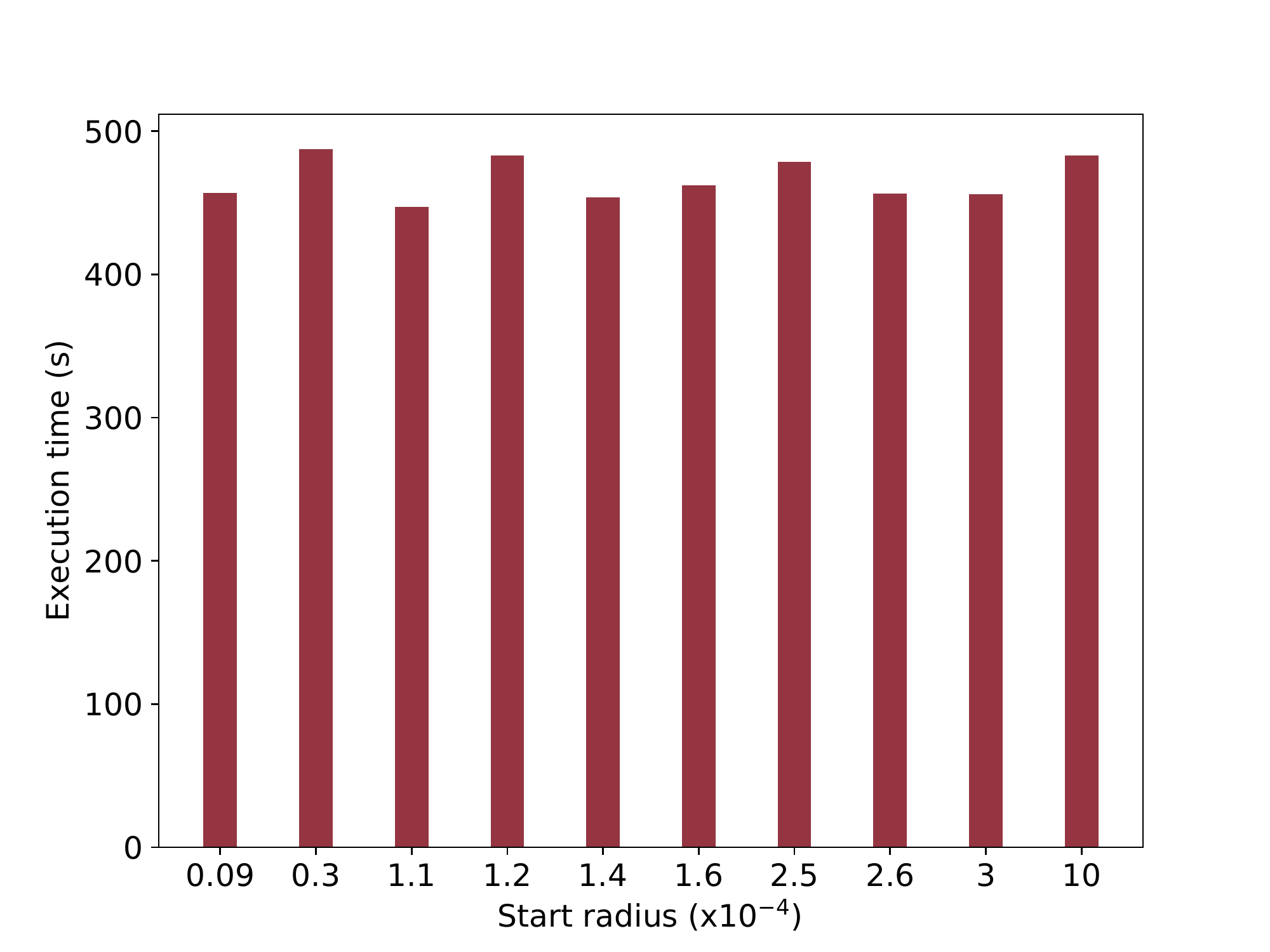}
    \vspace{-0.5em}
    \caption{Impact of start radius selection}
    \label{fig:start-radius}
    \vspace{-0.8em}
\end{figure}
In Section~\ref{sec:random_sample}, we proposed a method to choose a useful start radius for sphere expansion in the first stage of the {\tt TrueKNN} algorithm. In particular, we decided to always choose the minimum distance between the 4 nearest neighbors of 100 points as our starting radius. In this section, we show that this approach typically produces useful start radii.

We ran Algorithm~\ref{alg:random-sample} multiple times to generate different start radii for {\em 400K} points of the Porto dataset. Fig~\ref{fig:start-radius} shows the execution time (in seconds) of TrueKNN, where $k=\sqrt{DatasetSize}$, for the various start radii. In most cases, we found that there was an insignificant difference in the execution times for the different start radii. We found similar results on the other datasets and dataset sizes.


\subsection{Impact of outliers}
From our analysis in Section~\ref{sec:eval-runtime-analysis}, we see that using larger radii for neighbor search results in higher execution times even for a few query points. Since our baseline uses this large radius to query {\em all} points, it is unsurprising that it is much slower than TrueKNN.

In this section, we perform an experiment where we only identify neighbors upto the $99^{th}$ percentile distance to eliminate the influence of these outliers. We also create a synthetic dataset of points uniformly distributed over [0,1] to study the impact of outliers. In both experiments, we find that TrueKNN almost always outperforms the baseline. 

\subsubsection{$99^{th}$ Percentile} \label{sec:eval-99th}
We calculated the $99^{th}$ percentile search radius by computing the distance between query points and their neighbors and selecting the $99^{th}$ percentile distance. We perform this experiment to show that our approach does not unfairly benefit from the presence of outliers in the dataset. As the $99^{th}$ percentile radius is much smaller, the baseline greatly benefits from performing fewer intersection tests.

\begin{figure}[ht]
    \centering
    \includegraphics[width=0.45\textwidth]{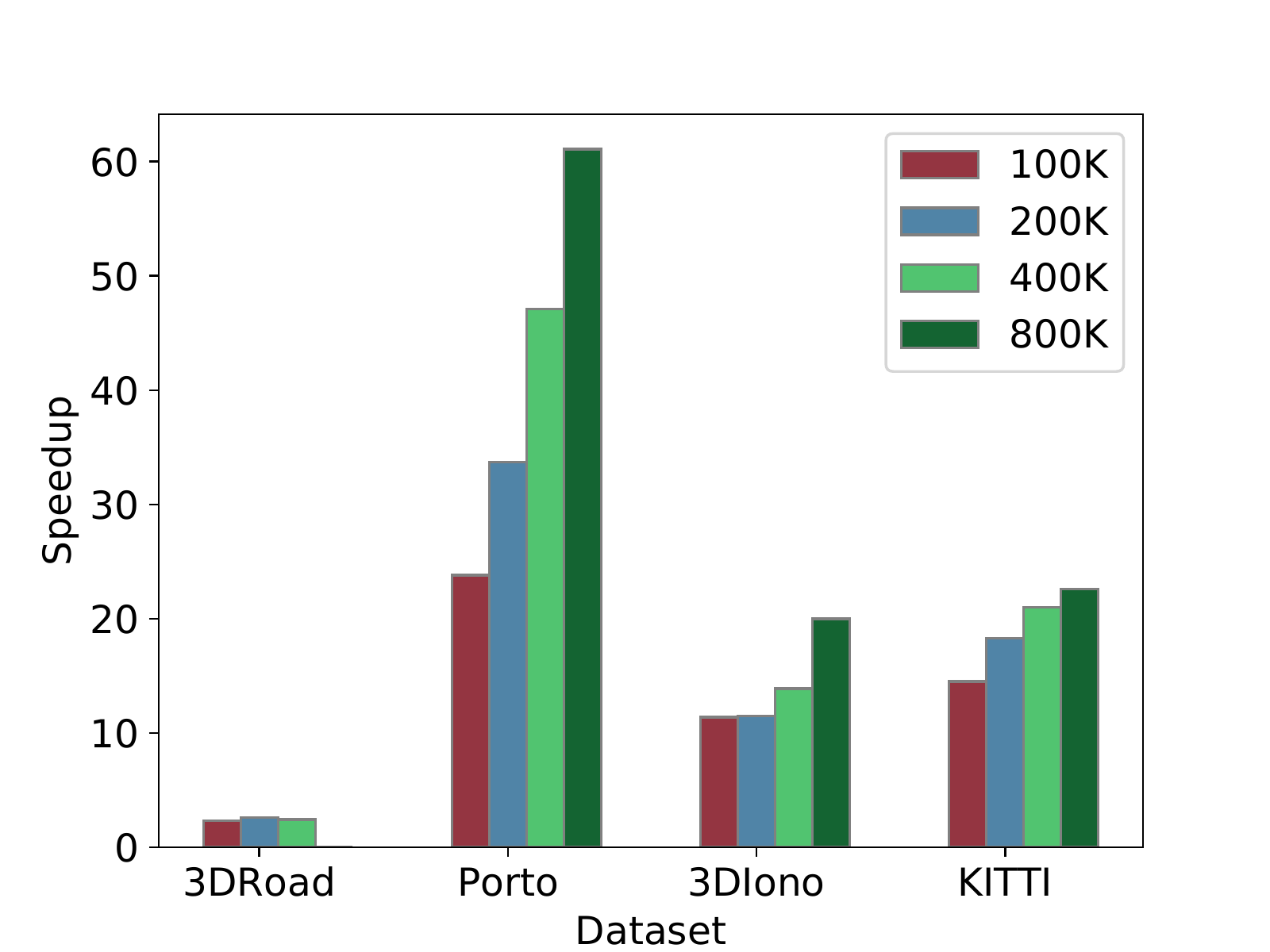}
    \vspace{-0.5em}
    \caption{TrueKNN's speedup compared to baseline for $99^{th}$ percentile neighbor search on different datasets}
    \label{fig:99th-percentile}
    \vspace{-0.8em}
\end{figure}

We evaluated performance on the Porto, 3DIono, and KITTI datasets, with the dataset size varying between {\em 100K} and {\em 800K} and $k = \sqrt{DatasetSize}$. We modified TrueKNN to terminate the execution when we reached the $99^{th}$ percentile radius. We note that this is only a thought experiment and that it is not possible to know the $99^{th}$ percentile without {\em actually} computing the neighbors of each query point. From Fig~\ref{fig:99th-percentile}, it is evident that TrueKNN still outperformed the baseline in all cases. This showed us that even with a drastic reduction in the search radius for the baseline ($\approx 30x$ on average), our approach still manages to reduce the number of software intersection tests. {\em In fact, our original TrueKNN, which identifies all neighbors, is also faster than the $99^{th}$ percentile-modified baseline in all cases!}

\subsubsection{Uniformly Distributed Dataset}
We created a synthetic dataset of 1M points that are uniformly distributed on [0,1] to see whether our approach would work well in the absence of blatant outliers. We studied the effects of varying dataset size and also repeated our $99^{th}$ percentile experiment on this dataset for $k = \sqrt{DatasetSize}$.

In both experiments, TrueKNN outperformed the baseline, as shown in Table~\ref{table:unif-dist}. This was surprising since this would be the worst-case input to TrueKNN, due to the absence of outliers. Though our speedup margin is not as high as the previous experiments, we still manage to perform fewer intersection tests, resulting in speedups of up to 4.2x on the regular kNNS experiment and 1.7x on the $99^{th}$ percentile experiment.
\begin{table}[h]
\centering
    \begin{tabular}{lcccl} 
    \toprule
       Dataset size  & kNNS & $99^{th}$ percentile kNNS \\
       \midrule
       100K & 3.5x & 1.5x  \\

       200K & 3.25x & 1.23x  \\

       400K &  4.28x & 1.7x  \\

       800K & 4.15x & 1.78x \\
       \bottomrule
   \end{tabular}
   \caption{TrueKNN's speedup over baseline on kNNS problems for UniformDist Dataset}
\label{table:unif-dist}
\end{table}


\section{Discussion}
In this section, we discuss TrueKNN's ability to accelerate {\em fixed-radius} kNNS and look at the hardware limitations that affect the efficiency of RT-accelerated general purpose applications.

\subsection{Fixed-radius kNNS}
Fixed-radius kNNS is a variant of kNNS, where the search space for neighbors is restricted to a particular radius.
From our experiments in Section~\ref{sec:eval}, we believe that, in addition to unbounded kNNS, our approach could {\em also} work well for fixed-radius kNNS applications. 

In the $99^{th}$ percentile experiment in Section~\ref{sec:eval-99th}, which is a  fixed-radius kNNS problem, we found that we were consistently faster than the fixed-radius baseline. This observation gives us an important result: our multi-round approach can be effectively used even for {\em fixed} radius neighbor searches, as our approach minimizes the number of intersection tests performed. For large datasets and large {\em k} values, TrueKNN's overhead costs are amortized, as we will see in Section~\ref{sec:disc-overhead}.
\begin{figure}[ht]
    \centering
    \includegraphics[width=0.45\textwidth]{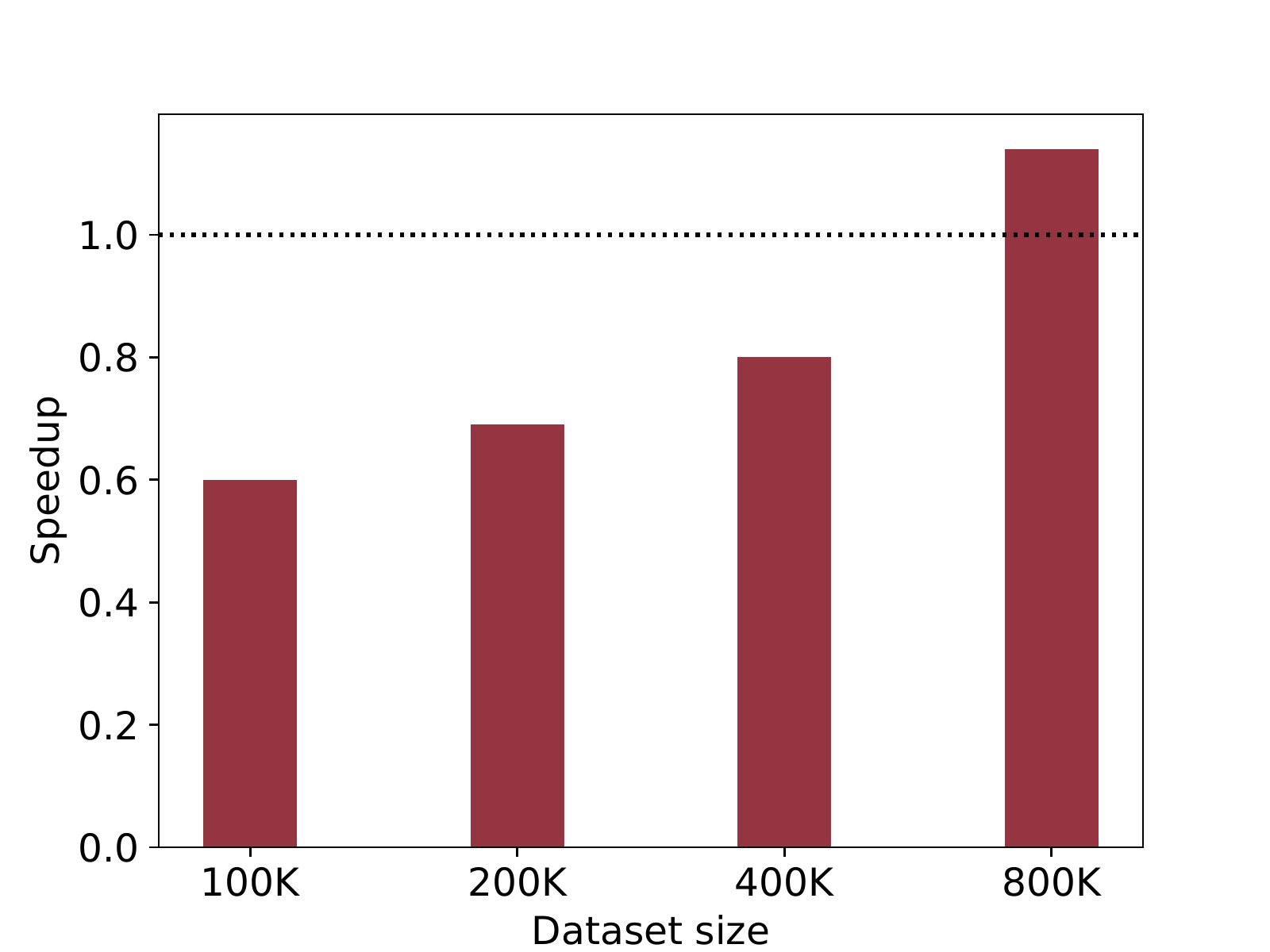}
    \vspace{-0.5em}
    \caption{TrueKNN's speedup compared to baseline for $99^{th}$ percentile neighbor search on 3DIono dataset}
    \label{fig:3diono-99th-k5}
    \vspace{-0.8em}
\end{figure}
For smaller datasets and {\em k} values, we find that TrueKNN is sometimes slower than the baseline as these overhead costs are not amortized. We experimented with {\em 100K}-{\em 800K} points from the Porto, 3DRoad, 3DIono and KITTI datasets and set $k = 5$. Though we were consistently faster on the  Porto, 3DRoad and KITTI datasets, we were upto 1.6x slower on the 3DIono dataset, as shown in Fig~\ref{fig:3diono-99th-k5}. This is mainly due to the overhead of switching between host and device contexts and the cost is not amortized across the iterations needed to find all neighbors. 

\subsection{Limitations of RT Hardware and API}
The fundamental limitation of using the RT hardware is the restriction to 3D datasets. As the hardware was created to accelerate graphics rendering, it was designed to work {\em only} with 3D data. We can get the hardware to work for 2D and 1D datasets by setting the corresponding dimensions to 0. For example, we set the z-dimension to 0 for 2D datasets, allowing the hardware to treat it as a 3D dataset.

Though it is not possible to express higher dimensional datasets using this hardware, we can use dimensionality reduction techniques such as Principal Component Analysis (PCA) \cite{PCA-Hotelling1933AnalysisOA}, Linear Discriminant Analysis (LDA) and Generalized Discriminant Analysis (GDA) \cite{GDA} to reduce the multi-dimensional dataset to just 3 dimensions. We also note that there are, in fact, many important 2D and 3D datasets, such as point clouds, geospatial data and geometries. These datasets are widely used in applications such as clustering and surface normal computation, both of which use kNNS as a subroutine.

\subsubsection{Optix Overhead Costs} \label{sec:disc-overhead}
As the OWL API was intended to serve ray tracing applications, its setup is not ideal for general-purpose computations. We notice this in particular when we implement our multi-stage TrueKNN algorithm (Section~\ref{sec:design-trueknn}). At the end of each round, we re-fit the bounding boxes around the spheres to expand our neighbor search space. For this operation, we need to switch the context from the device to the host to increment the size of the bounding boxes in the BVH. We then call the RayGen kernel (Fig~\ref{fig:optix-api}) to begin the ray casting process, which requires a context switch from host to device. As the Optix API does not allow the BVH re-fit to happen on the device side, we incur this cost every round. 

\section{Related Work}
\paragraph{Leveraging RT cores for non-Ray-Tracing Applications}
The idea of using RT cores to accelerate applications other than ray tracing was first introduced by Wald\etal~\cite{wald19}. They formulated the problem of identifying a point's location in a tetrahedral mesh as a ray tracing problem by declaring the meshes as 3D objects in a scene and tracing rays originating at the query point. They show how leveraging both hardware-accelerated BVH traversal and ray-triangle intersection tests resulted in up to 6.5x speedup over other CUDA implementations. Morrical\etal used RT cores to accelerate the problem of finding a point's location in unstructured elements with both
planar and bilinear faces~\cite{Morrical2020AcceleratingUM} and rendering of unstructured meshes~\cite{Morrical2019EfficientSS}. Zellmann\etal proposed a mapping of the fixed-radius nearest neighbor query to ray tracing queries by expanding spheres over points in the dataset and launching a small ray to record intersections~\cite{force-directed-graph}. They used the nearest neighbor query as a subroutine for the Spring Embedders force-directed graph drawing algorithm and show performance improvement between 4x to 13x over purely CUDA-based implementations. Evangelou\etal used the nearest neighbor mapping to solve the {\em k}-Nearest Neighbors problem, which returns the {\em k} closest neighbors of a point, and also perform photon mapping~\cite{Evangelou2021RadiusSearch}. Zhu proposed z-order sorting and query partitioning of input data points to reduce control flow and memory divergence in RT-accelerated neighbor searches~\cite{rtnn}.
\paragraph{$k$-Nearest Neighbor Search}
The $k$-Nearest Neighbor Search (kNNS) algorithm is used to find similarities within the feature space of input vectors. Researchers proposed using index structures to optimize the neighbor search, but soon found that this approach was useful only when the dataset had less than 10 dimensions\cite{knn-high-dim}. Nagarkar\etal provides an overview of popular indexing techniques for lower and higher dimensional data~\cite{knns-survey}. Some of the techniques used in lower dimensions include M Tree~\cite{m-tree}, R/R* tree~\cite{r-tree}, and k-d Tree~\cite{kd-tree}. Hashing and quantization algorithms were used to find Approximate Nearest Neighbors (ANN) in higher dimensions~\cite{hashing-ann,ann}.

Researchers have worked on leveraging GPU acceleration for KNNS as it is often the computational bottleneck in applications. Qui\etal implemented kNNS on the GPU to accelerate point cloud registration and show that it is 88x faster than its CPU counterpart~\cite{point-cloud-gpu-knn}. Johnson\etal created a suite of GPU-accelerated approximate neighbor searches called FAISS that used quantization optimizations~\cite{faiss-johnson2019billion}. Wieschollek\etal also uses a quantization-based approach in the form of Product Quantization Trees (PQT) to accelerate ANN~\cite{ann-gpu} on GPUs.

\section{Conclusion}
In this work, we implemented TrueKNN, the first RT-accelerated $k$-Nearest Neighbor Search algorithm that {\em does not} restrict the neighbor search space to a pre-defined fixed radius. TrueKNN uses an iterative approach where we initially sample the input dataset to guess a good search radius, and then incrementally increase the search space such that each point finds its $k$ nearest neighbors. We found that TrueKNN was orders of magnitude faster than existing algorithms on the {\em unbounded} neighbor search task and significantly faster even on the {\em fixed-radius} neighbor search task.

\begin{acks}
We thank the anonymous ICS reviewers for their valuable feedback. We are grateful to Kirshanthan Sundararajah for his comments that helped improve the paper. This work was funded by NSF grants CCF-1908504, CCF-1919197 and CCF-2216978.
\end{acks}

\bibliographystyle{ACM-Reference-Format}
\bibliography{ref}

\end{document}